\documentclass[10pt,twocolumn,letterpaper]{article}

\usepackage{cvpr}
\usepackage{times}
\usepackage{epsfig}
\usepackage{graphicx}
\usepackage{amsmath}
\usepackage{amssymb}
\usepackage{csquotes}
\usepackage{algorithm}
\usepackage{algorithmic}

 \cvprfinalcopy 

\ifcvprfinal\fi
\begin{document}

\title{FFNet: Video Fast-Forwarding via Reinforcement Learning}

\author{Shuyue Lan \textsuperscript{1}\\
\and
Rameswar Panda \textsuperscript{2}\\
\and
Qi Zhu \textsuperscript{1}\\
\and
Amit K. Roy-Chowdhury \textsuperscript{2}\\
\and
 \textsuperscript{1}Northwestern University. 
{\tt \{slan@u., qzhu@\}northwestern.edu}
\and
 \textsuperscript{2}University of California, Riverside.
{\tt \{rpand002@, amitrc@ece.\}ucr.edu}
}

\newenvironment{myitemize}{\begin{list}{$\bullet$}
		{\setlength{\topsep}{1mm}
			\setlength{\itemsep}{0.25mm}
			\setlength{\parsep}{0.25mm}
			\setlength{\itemindent}{0mm}
			\setlength{\partopsep}{0mm}
			\setlength{\labelwidth}{15mm}
			\setlength{\leftmargin}{4mm}}}{\end{list}}

\maketitle

\begin{abstract}
	
	For many applications with limited computation, communication, storage and energy resources, there is an imperative need of computer vision methods that could select an informative subset of the input video for efficient processing at or near real time. In the literature, there are two relevant groups of approaches: generating a ``trailer'' for a video or fast-forwarding while watching/processing the video. The first group is supported by video summarization techniques, which require processing of the entire video to select an important subset for showing to users. In the second group, current fast-forwarding methods depend on either manual control or automatic adaptation of playback speed, which often do not present an accurate representation and may still require processing of every frame. In this paper, we introduce FastForwardNet (FFNet), a reinforcement learning agent that gets inspiration from video summarization and does fast-forwarding differently. It is an online framework that automatically fast-forwards a video and presents a representative subset of frames to users on the fly. It does not require processing the entire video, but just the portion that is selected by the fast-forward agent, which makes the process very computationally efficient. The online nature of our proposed method also enables the users to begin fast-forwarding at any point of the video. Experiments on two real-world datasets demonstrate that our method can provide better representation of the input video (about $6\%$-$20\%$ improvement on coverage of important frames) with much less processing requirement (more than $80\%$ reduction in the number of frames processed).
	
\end{abstract}

\section{Introduction}

\begin{figure}
	\begin{center}
		\includegraphics[width=0.9\linewidth]{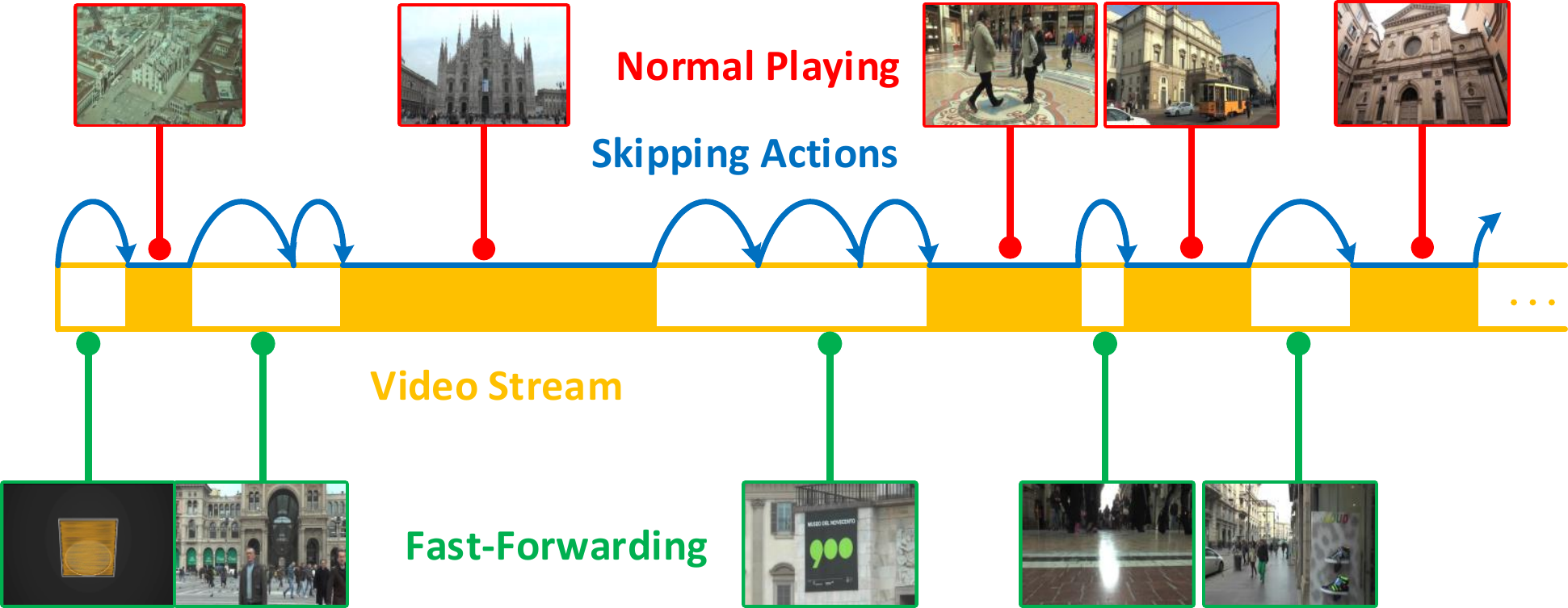}
	\end{center}
	\caption{\textbf{Overview of Our Proposed Method.} Given a video stream, our FFNet decides which frame to process next and presents it to users while \emph{skipping} the irrelevant frames in an online manner. Top-row shows the representative frames in the normal playing portion and bottom-row shows the irrelevant frames in the fast-forwarding portion. Best viewed in color.
	}
	\label{fig:motivation}
	\vspace{-6pt}
\end{figure}

Leveraging video input has become increasingly important in many intelligent Internet-of-Things (IoT) applications, such as environment monitoring, search and rescue, smart surveillance, and wearable devices. In these systems, large amount of video needs to be collected and processed by users (human operators or autonomous agents), either locally or remotely through network transmission (or a combination of both). For better system performance, the processing often needs to be done at or near real time. On the other hand, the local nodes/devices typically have limited computation and storage capability and often run on batteries, while the communication network is constrained by bandwidth, speed and reliability~\cite{akyildiz2007survey,ma2013survey,singh2014survey}. Such discrepancy presents an urgent need for new vision methods that can automatically select an informative subset of the input video for processing, to reduce computation, communication and storage requirements and to conserve energy. 

In the relevant literature, with ever expanding volume of video data, there is significant interest in video summarization techniques, which compute a short and informative subset of the original video for human consumption or further processing~\cite{elhamifar2017online,gygli2015video,panda2017weakly,zhang2016summary,zhang2016video,zhao2014quasi}. However, these techniques require the processing of entire video and often take a long time to generate the subset. There are also video fast-forwarding techniques where the playback speed of the video is adjusted to meet the needs of users~\cite{cheng2009smartplayer,halperin2017egosampling,joshi2015real,petrovic2005adaptive,poleg2015egosampling,ramos2016fast,silva2016towards}, but they often do not present an accurate representation and may still require processing of the entire video. Both types of approaches are not suitable for the resource-limited and time-critical systems we discussed above. To address this problem, we started by asking the following question: \emph{Is it possible to develop a method for fast-forwarding through a video that is computationally efficient, causal, online and results in informative segments which can be validated through statistical evaluation and user experience?}

In this paper, we introduce FastForwardNet (FFNet), a reinforcement learning agent that gets inspiration from video summarization and does fast-forwarding differently from the state-of-the-art methods. It has an online framework that automatically fast-forwards a video and presents a selected subset of frames to users on the fly (see Fig.~\ref{fig:motivation} for an example). The fast-forward agent does not require the processing of entire video. This makes the process very computationally efficient, and communicationally efficient if the video (subset) needs to be transmitted over the network for remote processing. The online nature of our proposed FFNet enables the users to begin fast-forwarding at any point when watching/processing videos. The causal nature of our FFNet ensures that it can work even as the video subset is being generated.

To summarize, the key advantage of our approach is that it automatically selects the most important frames \textbf{\textit{without}} processing or even obtaining the entire video. Such capability can significantly reduce resource requirements and lower energy consumption, and is particularly important for resource-constrained and time-critical systems. The main technical contributions of this paper are as follows.

\vspace{1mm}

\noindent (1) We formulate video fast-forwarding as a Markov decision process (MDP), and propose FFNet for fast-forwarding a potentially very long video while presenting its important and interesting content on the fly.

\vspace{1mm}

\noindent (2) We propose an online framework to deal with incremental observations without requiring to store and process the entire video. At any point of the video, our approach can jump to potentially important future frames based on analysis of past frames that had been selected.

\vspace{1mm}

\noindent (3)  We demonstrate the effectiveness of our proposed FFNet on two standard challenging video summarization datasets, Tour20~\cite{panda2017diversity} and TVSum~\cite{song2015tvsum}, achieving real-time speed on all tested videos.

\section{Related Work}

Our work relates to three major research directions: video fast-forward, video summarization and reinforcement learning. Here, we focus on some representative methods that are closely related to our work.

\textbf{Video Fast-Forward.} Video fast-forward methods are typically used when users are losing patience to watch the entire video. Most commercial video players offer manual control on the playback speed, e.g., Apple QuickTime Player offers 2, 5 and 10 multi-speed fast-forward. 

Current automatic fast-forward approaches mostly focus on adapting the playback speed based on either the similarity of each candidate clip to the query clip~\cite{petrovic2005adaptive} or the motion activity patterns present in a video~\cite{cheng2009smartplayer,peker2003extended,peker2001constant}. Some recent works use mutual information between frames to describe the fast-forward policy~\cite{jiang2010new,jiang2011smart}, or use shortest path distance over the graph that is constructed with semantic information extracted from frames~\cite{ramos2016fast,silva2016towards}. This family of methods is most relevant to our goal. A similar family of work (hyperlapse)~\cite{poleg2015egosampling,halperin2017egosampling,joshi2015real} aiming at speed-up and smoothing has also been developed for creating fast-forwarded videos. In contrast to these prior works, we develop a deep reinforcement learning strategy for the fast-forward policy. Our proposed framework (FFNet) is an on-line and causal system that does not need the entire video to get the fast-forward policy, making it very efficient in terms of computation, communication and storage needs. 

\textbf{Video Summarization.} The goal of video summarization is to produce a compact summary that contains the most important parts of a video. Much progress has been made to summarize a video using either supervised learning based on video-summary pairs~\cite{gong2014diverse,gygli2015video,zhang2016summary,zhang2016video,Category2014,panda2017weakly} or unsupervised approaches based on low-level visual indices~\cite{elhamifar2012see,gygli2014creating,Att2005,Top2014} (see reviews~\cite{money2008video,truong2007video}). Leveraging crawled web images or videos is another recent trend for video summarization~\cite{khosla2013large,Joint2014,song2015tvsum,panda2017collaborative}. Closely related to video summarization, the authors in~\cite{sigurdsson2016learning} develop a framework for creating storylines from photo albums.

Most relevant to our approach is the work of online video summarization, which compiles the most salient and informative portion of a video by automatically scanning through the video stream, in an online fashion, to remove repetitive and uninteresting content. Various strategies have been studied, including Gaussian mixture model~\cite{ou2014low}, online dictionary learning~\cite{zhao2014quasi}, and submodular optimization~\cite{elhamifar2017online}. Our approach significantly differs from these methods in that it only processes a subset of frames instead of the entire video. To the best of our knowledge, this is the first work to address video fast-forwarding in generating an informative summary from a video.

\textbf{Reinforcement Learning.}  Apart from the recent success in playing Go games and Atari~\cite{mnih2015human,silver2016mastering}, deep reinforcement learning(DRL) has also achieved promising performance in several vision tasks, such as object detection~\cite{Mathe_2016_CVPR}, visual tracking~\cite{Yun_2017_CVPR}, pose estimation~\cite{Krull_2017_CVPR} and image captioning~\cite{Ren_2017_CVPR}. \cite{yeung2016end} employs a computationally intensive reinforcement learning strategy for action detection in short video clips. 
In contrast to \cite{yeung2016end}, our framework is an online and causal system that enables users to begin fast-forwarding at any point while watching videos (online) and can work even as summary is being generated (causal).
Markov Decision Process (MDP) has been widely used for several vision tasks. For example, in~\cite{su2016leaving}, the authors formulate a policy learning as MDP for activity recognition. Q-learning (a reinforcement learning method) is one way to solve MDP problems~\cite{watkins1992q}. In the proposed FFNet, we use a multi-layer neural network to represent the Q-value function, similar to~\cite{riedmiller2005neural,wei2017deep}. We are not aware of any prior work in reinforcement learning that deals with fast-forwarding while summarizing long duration videos.

\section{Methodology}

In this section, we provide the details of FFNet.  We start with an overview of our approach in Sec.~\ref{sec3.1}, present detailed formulations in Sec.~\ref{sec3.2} and Sec.~\ref{sec3.3}, and then explain the training algorithm in~\ref{sec3.4}.

\begin{figure*}
	\begin{center}
		\includegraphics[width=\linewidth]{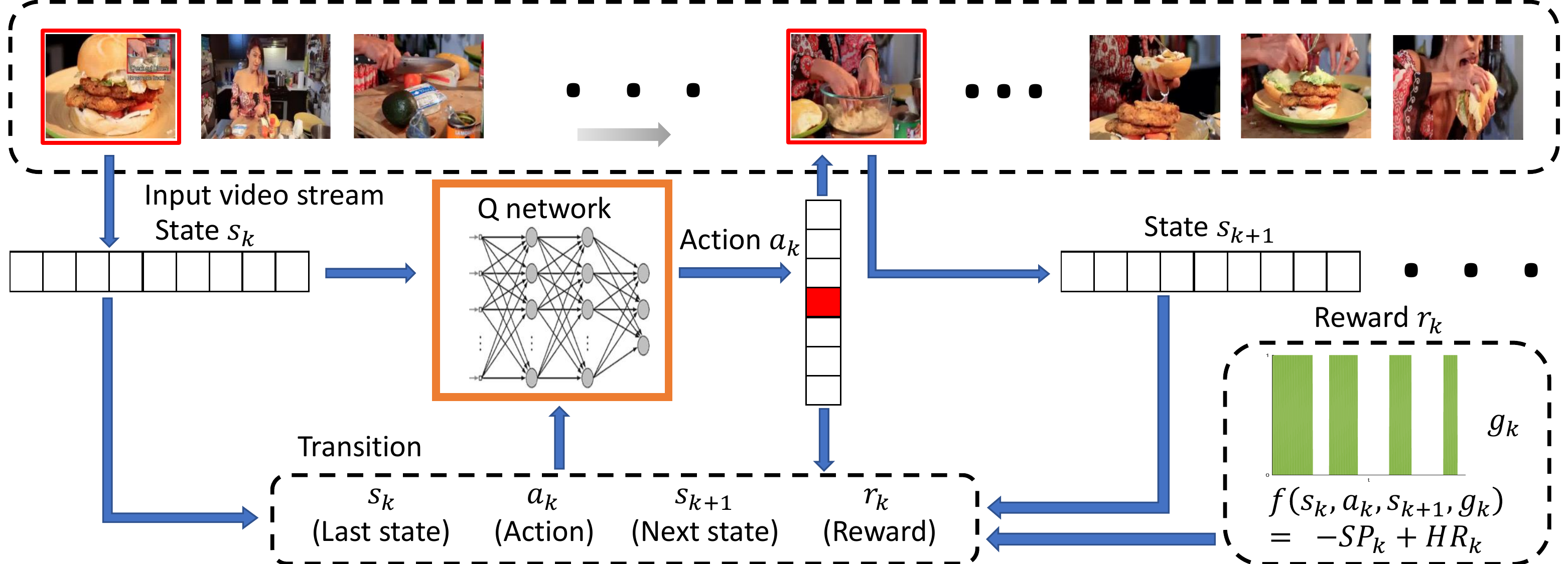}
	\end{center}
	\caption{\textbf{Model of our FFNet.} We learn a strategy for fast-forwarding videos. At each time step $k$ we use the Q network to select the action $a_k$, i.e., the number of frames to skip next. The state $s_{k+1}$ is updated with the frame it jumps to. Then, the reward $r_k$ for action $a_k$ in state $s_k$ is computed with the interval annotation $g_k$. A transition  in a quadruple form ($s_k$ $a_k$ $s_{k+1}$ $r_k$) is used to update the Q network.}
	\label{model}
\end{figure*}

\subsection{Solution Overview} \label{sec3.1}

Our goal is to fast-forward a long video sequence by skipping unimportant frames in an online and real-time fashion (see Fig.~\ref{fig:motivation}). 
Given the current frame being processed, the goal of FFNet is to decide the number of frames to skip next. Those frames within the skipping interval will not be processed.
Then, the video frames we present to users include the frames processed by FFNet and their neighboring windows (which are not processed).

We formulate the above fast-forwarding problem using a Markov decision process (MDP) and develop our FFNet as a reinforcement learning agent, i.e., a Q-learning agent that learns a policy to skip unimportant frames and present the important ones to users. During test time, given a raw video, fast-forwarding is a sequential process. At each step $k = 1,..., K$ of an episode, we process the current frame, decide how many future frames to skip, and jump to the frame after the skipped ones for next processing. We present the processed frames and their neighboring ones (with processed frames as window centroids) to users as important subsets of the video.

\subsection{MDP Formulation for FFNet} \label{sec3.2}

We consider fast-forwarding as a control problem that can be formulated as an MDP with the following elements.
\noindent\textbf{State:}  A state $s_k$ describes the current environment at the k-th step of the episode.  Given a video sequence, we consider a single frame as a state, defined in terms of the extracted feature vector of the current frame. 

\noindent\textbf{Action:} An action $a_k$ is performed at step k by the system and leads to an update of the state.  We define a discrete set of possible actions $A=\left\{a^1, a^2, ..., a^M\right\}$, which represents the possible numbers of frames to skip.

\noindent\textbf{Reward:} An immediate reward $r_k = r(s_k, a_k, s_{k+1})$ is received by the system when it transits from one state $s_k$ to another state $s_{k+1}$ after taking action $a_k$ (Sec.~\ref{sec3.3}).  

The accumulated reward is then defined as 
\begin{equation} \label{totalreward}
	R = \sum_{k}^{} \gamma^{k-1}r_k = \sum_{k}^{} \gamma^{k-1}r(s_k, a_k, s_{k+1})
\end{equation}
where $\gamma \in [0, 1] $ denotes the discount factor for the rewards in the future.

\noindent \textbf{Policy:} The policy $\pi$ determines the action to be chosen in every state visited by the system, i.e., it selects the action that maximizes the expected accumulated reward for current and future actions as
\begin{equation}
	\pi(s_k) = arg \max_{a} E[R|s_k, a, \pi]
\end{equation}

In this case, the policy in FFNet decides how many frames to skip when the system is at certain frame (state).

\subsection{Design of the Immediate Reward} \label{sec3.3}

In this part, we introduce the definition of the immediate reward $r_k$ for $a_k$ in state $s_k$. For a raw video available in the training set, we assume each frame $i$ has a binary label $l(i)$. $l(i) = 1$ indicates that frame $i$ is an important frame, and $l(i) = 0$ means it is an unimportant one.

Given a video and its labels, we define the immediate reward as follows:
\begin{equation} \label{reward}
	r_k =   - SP_k +HR_k
\end{equation}

The immediate reward consists of two parts that model the ``skip'' penalty (SP) and the ``hit'' reward (HR), as explained below.

First, $SP_k$ in Eqn.(\ref{reward}) defines the penalty for skipping the interval $t_k$ in step $k$:
\begin{equation}
	SP_k =  \frac{\sum_ {i \in t_k}^{ }\textbf{1}(l(i) = 1)}{T} - \beta \frac{\sum_{i \in t_k}^{ } \textbf{1}(l(i) = 0)}{T}
\end{equation}
where $\textbf{1}(\cdot)$ is an indicator function that equals to 1 if the condition holds. T is the largest number of frames we may skip, taken as a normalized term. $\beta \in [0,1]$ is a trade-off factor between the penalty for skipping important frames and the reward for skipping unimportant frames.

Then, the second term $HR_k$ in Eqn.(\ref{reward}) defines the reward for jumping to an important frame or a position near an important frame.  To model this reward, we first transfer the one-frame label to a Gaussian distribution in a time window. More specifically, a frame $i$ will have a reward effect on the positions in its nearby window that is defined as 
\begin{equation}
	f_i(t) = \frac{1}{\sqrt{2 \pi \sigma^2}} exp(- \frac{(t-i)^2}{2 \sigma^2}), t \in [i-w, i+w]
\end{equation}
where $w$ controls the window size of the Gaussian distribution. In the experiment section, we set $\sigma = 1, w = 4$. The reason for this transfer is that the reward should be given if the agent jumps to a position that is close to the important frame. To some extent, it jumps to a potentially important area. Assume in time step $k$, the agent jumps to the $z_{th}$ frame in the original video. Based on the above definition, the $HR_k$ is computed as
\begin{equation}
	HR_k =  \sum_{i=z-w}^{z+w} \textbf{1}(l(i)=1) \cdot f_i(z)
\end{equation}

\subsection{Learning the Fast-Forwarding Policy} \label{sec3.4}

During the operation of FFNet, we want to maximize the accumulated reward $R$ in Eqn.(\ref{totalreward}). Our goal is to find an optimal policy $\pi^*$ that maps the state to the corresponding action to fulfill the requirement. With Q-learning, we evaluate the value of action $E[R|s,a,\pi]$ as $Q(s,a)$. In classical Q-learning method, the Q-value is updated by 
\begin{equation} \label{qupdate}
	\begin{split}
		Q_{k+1}(s_k,a_k) = & (1- \alpha)Q_k(s_k,a_k) \\
		& + \alpha(r_k + \gamma \max_{a_{k+1}}Q_k(s_{k+1}, a_{k+1}) )
	\end{split}
\end{equation}
where $\alpha \in (0,1]$ represents the learning rate during the training process.

In this problem, we have finite actions but infinite states. No direct assignment of Q-values can be made, thus we use the neural network to approximate the Q-value. The Q-function in this work is modeled by a similar multilayer perception (MLP) structure as in \cite{wei2017deep}. The input is the current state vector, and the output is a vector of the estimated Q-value for each action given the state. The optimal value of the accumulated reward in time step $k$ is achieved by taking action $a_k$ and represented as $Q^*(s_k,a_k)$, which can be calculated by Bellman equation in a recursive fashion:
\begin{equation} \label{bellman}
	Q^*(s_k,a_k) = r_k + \gamma \max_{a_{k+1}}Q^*(s_{k+1}, a_{k+1})
\end{equation}
where $\gamma$ is the same discount factor in the definition of the accumulated reward in Eqn. (\ref{totalreward}). Note that when using gradient descent, Eqn.(\ref{bellman}) is consistent with the Q-learning update equation Eqn. (\ref{qupdate}).

With the above update equation, we use the mean squared error between the target Q-value and the output of MLP as the loss function. During the training process, we apply $\epsilon$-greedy strategy to better explore the state space, which picks a random action with probability $\epsilon$ and the action that has $Q^*(s,a)$ with probability 1-$\epsilon$. 

The model of our FFNet is shown in Fig. \ref{model}. Given a video, the fast-forward agent starts from the first frame. The FFNet $Q$ is initialized with random parameters. For the current frame in time step $k$, we first extract the feature vector to get the state $s_k$. Based on the current $Q$ network and the state $s_k$, one action $a_k$ is chosen using the $\epsilon$-greedy strategy and the agent jumps to a new frame based on the action. Then the current state transits to $s_{k+1}$, i.e., the feature extracted from the new current frame. With the interval labels $g_k$ of the video, we compute the immediate reward $r_k$ for performing this action. The transition ($s_k$, $a_k$, $s_{k+1}$, $r_k$) then is sent to update the $Q$ network. More details about our training algorithm are presented in Algorithm \ref{alg1}.

\begin{algorithm}
	\caption{Training Algorithm for FFNet}
	\label{alg1}
	\begin{algorithmic}[1]
		\STATE \textbf{Input:} a set of videos $\left\lbrace V \right\rbrace $ and annotations  $\left\lbrace G \right\rbrace $
		\STATE \textbf{Output:} Q-value neural network $Q$
		\STATE Init\_MLP( ) $\rightarrow Q$
		\STATE Initialize: memory $M$ = [empty], $explore\_rate $ $\epsilon$ = 1
		
		\FOR{$i = 1$ to $N$}
		\STATE Training\_Video\_Selection ($V$, $G$) $\rightarrow v_i,g_i$
		\STATE $frame_{curr} = 0$
		\STATE Process($frame_{curr}$) $\rightarrow s_{curr}$
		
		\WHILE{$frame_{curr}<$Size($v_i$)}
		\STATE $a_{curr}=	
		\begin{cases}
		a^k \in A, k = random(n), & prob. = \epsilon \\
		arg \max Q(s_{curr},a'), & o.w.
		\end{cases}	$
		\STATE $frame_{next} = $ Action($a_{curr}$, $frame_{curr}$)
		\STATE Process($frame_{next}$) $\rightarrow s_{next}$
		\STATE $r =$ Reward($s_{curr}$, $a_{curr}$, $s_{next}$, $g_i$)
		\STATE $input = s_{curr}$ 
		\STATE $target=\begin{cases}
		r+\gamma \max_{a'} Q(s_{next},a'),& a=a_{curr} \\
		Q(s_{curr}, a), & o.w.
		\end{cases}$
		\STATE $(input, target) \rightarrow M$
		\STATE $ s_{next} \rightarrow s_{curr}$
		
		\IF{ $M>batch size$}
		\STATE Training($M$,$Q$) $\rightarrow Q$
		\STATE $\epsilon = \max (\epsilon-\vartriangle \epsilon, \epsilon_{min})$
		\STATE Empty($M$)
		\ENDIF

		\ENDWHILE
		\ENDFOR
	\end{algorithmic}	
\end{algorithm}

\section{Experiments}

In this section, we present extensive experiments and comparisons to demonstrate the effectiveness and efficiency of our proposed framework for fast-forwarding videos.

\subsection{Experimental Setup}

\textbf{Datasets.} We conduct experiments on two 
publicly available summarization datasets, namely Tour20~\cite{panda2017diversity} and TVSum~\cite{song2015tvsum}. 
Both datasets are very diverse. Tour20 consists of 140 videos of about 20 tourist attractions selected from the Tripadvisor travelers choice landmarks 2015 list. TVSUM contains 50 videos downloaded from YouTube in 10 categories, as defined in the TRECVid Multimedia Event Detection task. To the best of our knowledge, Tour20 is the largest publicly available summarization dataset with 140 videos totaling about 7 hours. Both datasets provide multiple user-annotated summaries for each video. For Tour20 dataset, we combine all three user summaries as human-created summary (labels for training). For TVSum dataset, we first average the frame-level importance scores to compute shot-level scores, and then select top 20\% shots for each video as human-created summary.

\textbf{Implementation Details.} Our FFNet is implemented using TensorFlow library on a Tesla K80 GPU. We use a 4-layer neural network to approximate the Q function. The discount factor $\gamma$ for the rewards in the future is set as 0.8. The exploration rate for Q-learning decays from 1 and stops at 0.1, with a rate of 0.00001. The memory size is set as 128 transitions. We train the Q network up to 1000 epochs for Tour20 and 800 epochs for TVSum.

\textbf{Performance Measures.} Similar to \cite{gygli2015video}, we report a coverage metric at video segment level, which measures how well the results of fast-forward methods cover the important frames in the ground truth obtained from human labeling. 
More specifically, a segment selected by a method is considered as true positive if the number of important frames it covers (based on the ground truth labels) exceeds a certain threshold called the \emph{hit number}. We evaluate on different hit numbers ranging from 1 to 20 throughout our experiments.

It is important to note that for the intelligent applications we target (e.g., smart surveillance, search and rescue), when measuring system performance, covering the important frames is more critical than skipping the unimportant ones, since such coverage determines the system's capability to identify important events and possibly react to them. 

\textbf{Compared Methods.}  We compare our approach with several methods that fall into two categories:
(1) offline processing methods including Microsoft Hyperlapse (MH)~\cite{joshi2015real}, Spectrual Clustering (SC)~\cite{von2007tutorial} and Sparse Modeling Representative Selection (SMRS)~\cite{elhamifar2012see}; and (2) online methods including LiveLight (LL)~\cite{zhao2014quasi} and Online Kmeans (OK)~\cite{arthur2007k}. Please see supplementary for more details.

\textbf{Experimental Settings.} We use Alexnet \cite{krizhevsky2012imagenet} fc7 features (4096-dimensional) to represent each video frame and tune the parameters in each method to have the best performance. For each method (including ours), we generate a subset of video frames that has the same length as in ground truth to make a fair comparison. We use the desktop version of Microsoft Hyperlapse (MH) to generate the subset videos in a 4x speed-up rate. For online k-means (OK) and spectral clustering (SC), we set the number of clusters to 20, as in~\cite{panda2017collaborative}. In LiveLight (LL), the dictionary is initialized as the first $10\%$ of segments in a video. For FFNet, we use an action space of 25, i.e., skipping from 1 to 25 frames and the trade-off factor $\beta$ is set to 0.8 throughout the experiments. For each dataset, we randomly select $80\%$ of videos for training and use the remaining $20\%$ for testing. We run 5 rounds of experiments and report the average performance.

\subsection{Coverage Evaluation}

\begin{figure}
	\begin{center}
		\includegraphics[width=\linewidth]{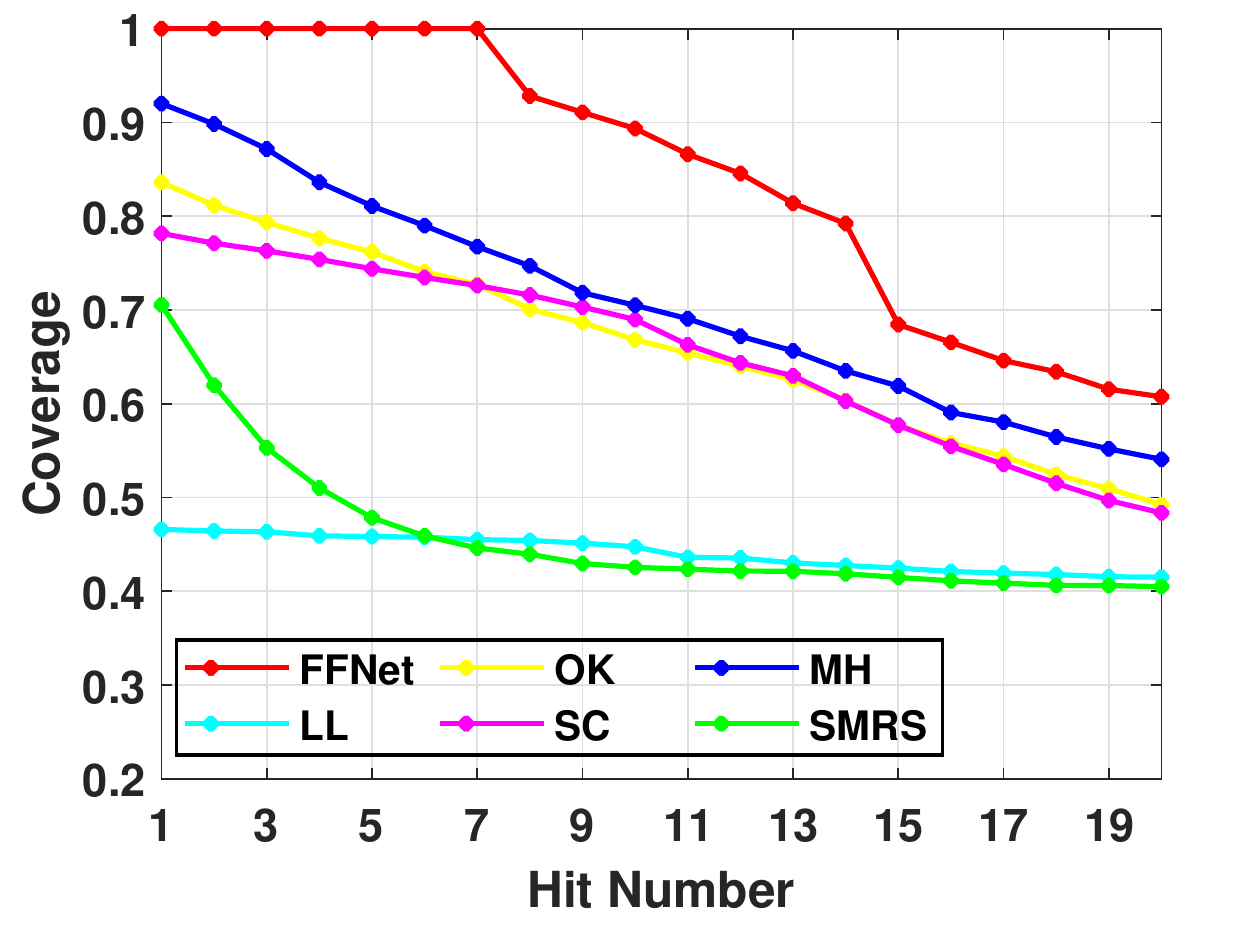} 
	\end{center}  \vspace{-4mm}
	\caption{\textbf{Segment-level coverage on Tour20 dataset with different hit number thresholds.} Our FFNet (red line on top) outperforms all other methods by a significant margin.}
	\label{tour20_r} 
\end{figure}

\begin{figure}
	\begin{center}
		\includegraphics[width=\linewidth]{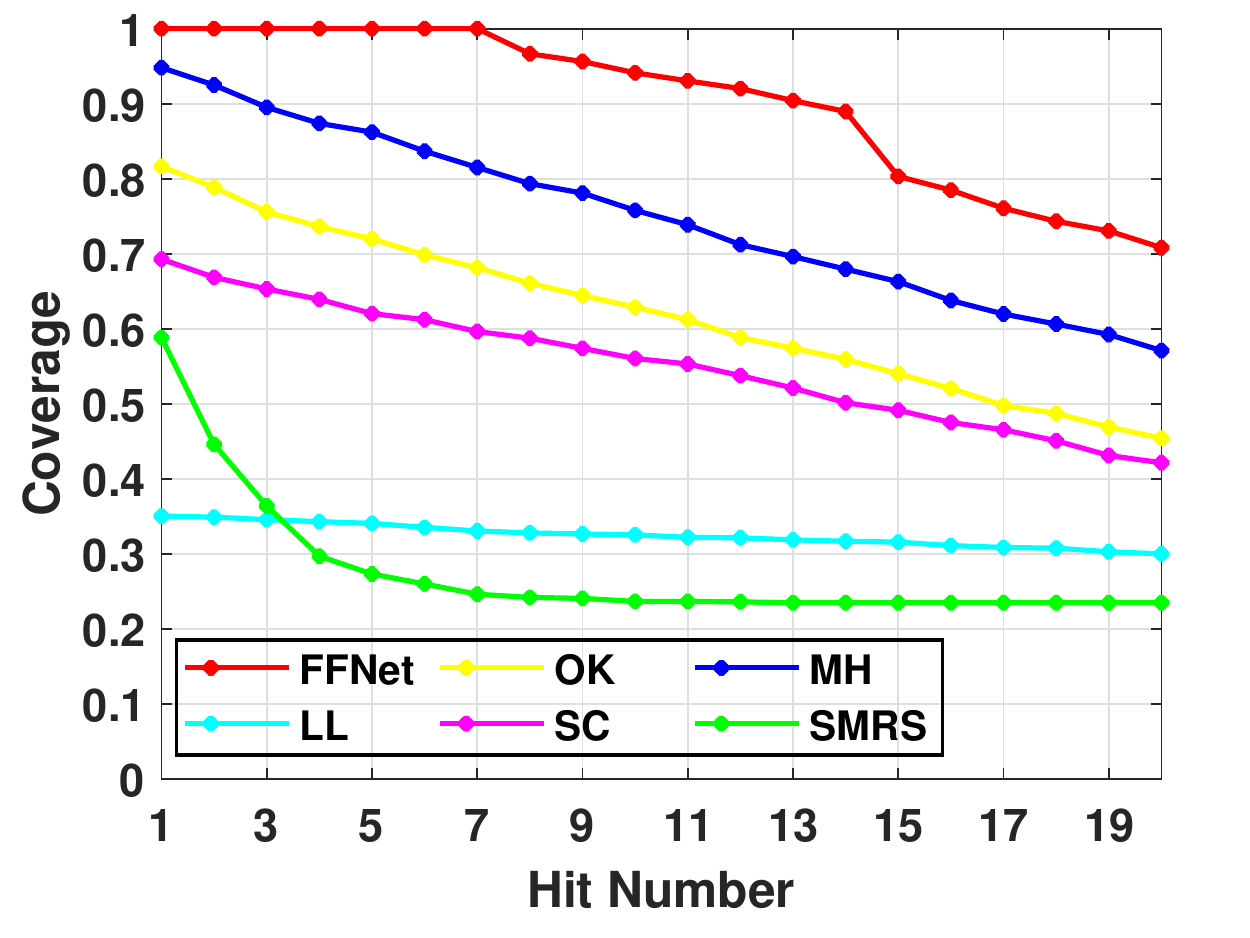}
	\end{center} \vspace{-4mm}
	\caption{\textbf{Segment-level coverage on TVSum dataset under different hit number thresholds.} Our FFNet (red line on top) outperforms all other methods by a significant margin.}
	\label{tvsum_r}
\end{figure}

Fig.~\ref{tour20_r} and Fig.~\ref{tvsum_r} show the mean segment-level coverage achieved by different methods on Tour20 and TVSum dataset, respectively. Each point in these figures represents the segment-level coverage achieved by an algorithm given certain hit number. For example, in Fig.~\ref{tour20_r}, our proposed FFNet achieves a segment-level coverage of about $90\%$ for a hit number of 10. This means that for about $90\%$ of the segments selected in ground truth (i.e., the important segments), at least 10 frames in each of them are selected/covered by FFNet. When the hit number is smaller or equal to 7, the coverage of FFNet is $100\%$, i.e., every important segment has at least 7 frames selected by FFNet.

When comparing FFNet with other methods in Fig.~\ref{tour20_r}, we have the following observations:
\begin{myitemize}
	\item For smaller hit numbers (say, under 10), our approach achieves excellent coverage ($90\%$ or above) and significantly outperforms all other methods (about $10\%$-$20\%$ better). This shows that \textbf{the subset selected by FFNet is able to provide more complete coverage of the important information throughout the video stream}.   
	\item As expected, the coverage of any method goes down with the increase of hit number requirement. Nevertheless, for larger hit numbers (say, 10-20), our approach FFNet still outperforms all other methods. This shows \textbf{its consistency in providing better coverage performance}.
\end{myitemize}

Similar result can be seen in Fig.~\ref{tvsum_r} for the TVSum dataset. Notice that for all methods (including ours), performance on Tour20 is not as good as on TVSum. We believe the difference is due to the fact that Tour20 dataset contains some videos capturing static objects and taken from a fixed camera. In this case, the state at each time step in our MDP is the same, which may confuse the Q-learning agent. 

\textbf{Comparison with State-of-the-Art Summarization Methods:} We additionally compare our FFNet with the state-of-the-art video summarization methods\cite{gygli2015video,panda2017collaborative,zhang2016video} and one supervised learning baseline (Sup) (implemented as regression) without reinforcement learning. Limited to space, we only present coverage at hit number of 10 in Table~\ref{coverage}. Note that we are only able to compare with \cite{zhang2016video} on the TVSum dataset as the pre-trained model is publicly available by the authors. Our approach outperforms all the baselines by a significant margin, showing that the summary selected by FFNet is able to provide more complete coverage of the important information throughout the video stream. Performance improvement over the Sup baseline shows the advantage of longer time horizon of the reinforcement learning policy in fast-forwarding videos.

\textbf{Qualitaive Results.} Fig.~\ref{tour20_ex} demonstrates a qualitative example from Tour20 dataset (see supplementary file for more of such examples). It clearly demonstrates that our FFNet is able to fast-forward through the unimportant parts and find the most important/relevant parts from a video, and is close to the ground truth (human-created summaries). At the top of Fig.~\ref{tour20_ex} are the representative video segments selected by our approach. The second row is the ground truth (GT). The remaining rows represent the segments selected by the other methods for the same video. At the beginning, our policy takes larger steps to skip frames that show only clouds without any interesting events. Once the roadside scenes (e.g., shopping area, walking tourists) start, the model begins to take small steps and presents most of the original segments. To summarize, we observe the following.

\begin{myitemize}
	\item For most of the important parts, our FFNet chooses not to skip and presents most of the original segments.
	\item For unimportant parts, FFNet takes larger jumping steps and smoothly skips frames. 
\end{myitemize}

\begin{figure*}
	\begin{center}
		\includegraphics[width=0.95\linewidth]{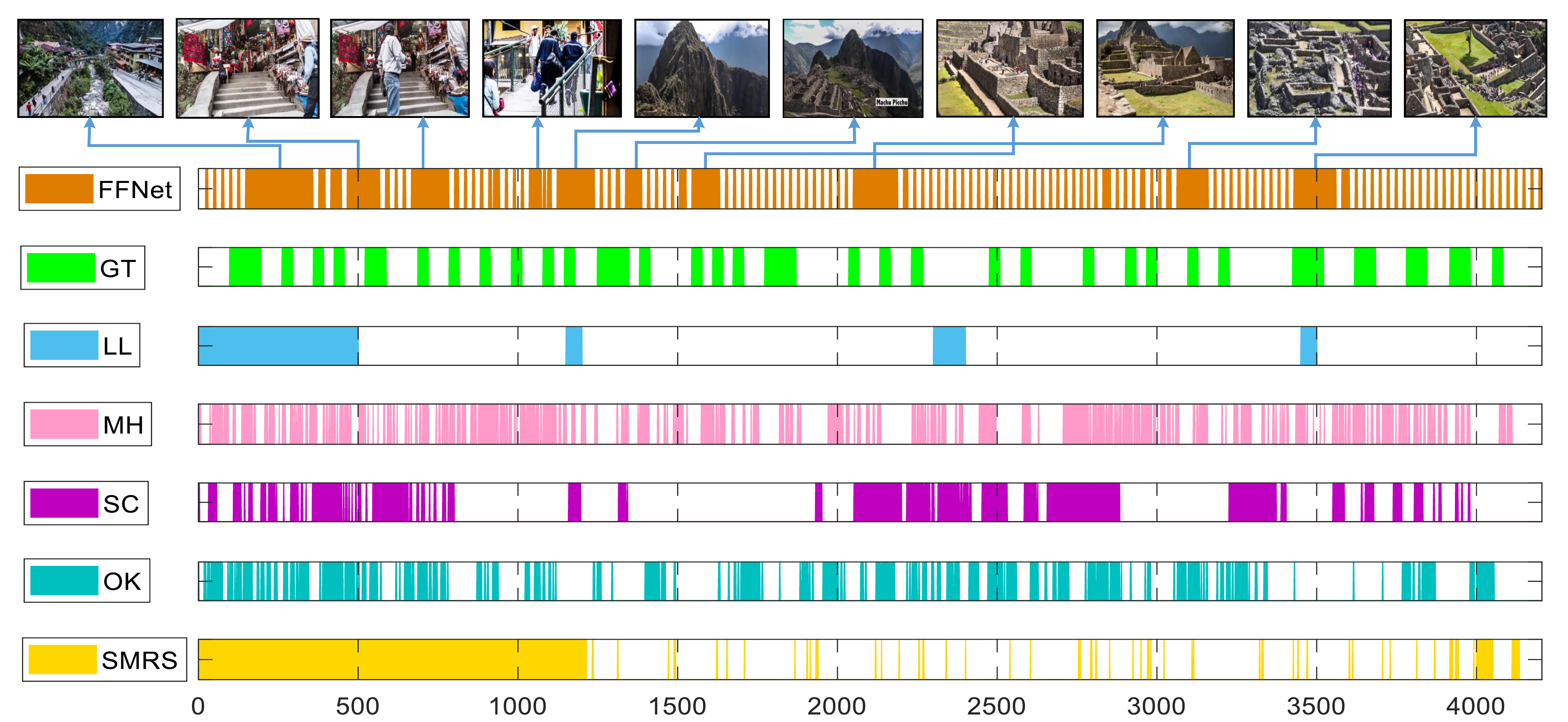}
	\end{center}
	\caption{\textbf{ Exemplar summaries generated while fastforwarding a video of Machu Picchu from the Tour20 dataset.} 
	The frames on top represent segments in our FFNet fast-forwarding result. The rows below illustrate the selected portions using different methods.
	The X-axis is the frame index over time. Notice that the segments selected by FFNet contains most of the important content labeled in the ground truth, including the roadside scene at the starting point, shopping area, walking tourists, different angles of the natural environment near the attraction, and the main citadel with zoom-in and zoom-out views. Figure is best viewed in color.}
	\label{tour20_ex} 
\end{figure*}

\begin{table}
	\begin{center}
		\begin{tabular}{|c|c|c|c|c|c|}
			\hline
			Dataset & \cite{gygli2015video} & \cite{panda2017collaborative} & \cite{zhang2016video} & Sup & FFNet  \\
			\hline\hline
			Tour20 & 0.754 & 0.826 & - & 0.685 & 0.893 \\
			\hline
			TVSum & 0.738 & 0.877 &0.553 & 0.526 & 0.941 \\
			\hline
		\end{tabular}
	\end{center}
	\caption{\textbf{Coverage achieved by different methods at hit number 10.}  \enquote{Sup} represents the supervised learning baseline with supervision being the \# of frames to jump to the next informative frame as per groundtruth. Our approach performs the best.}
	\label{coverage} \vspace{-2mm}
\end{table}

\begin{figure}
	\begin{center}
		\includegraphics[width=0.9\linewidth]{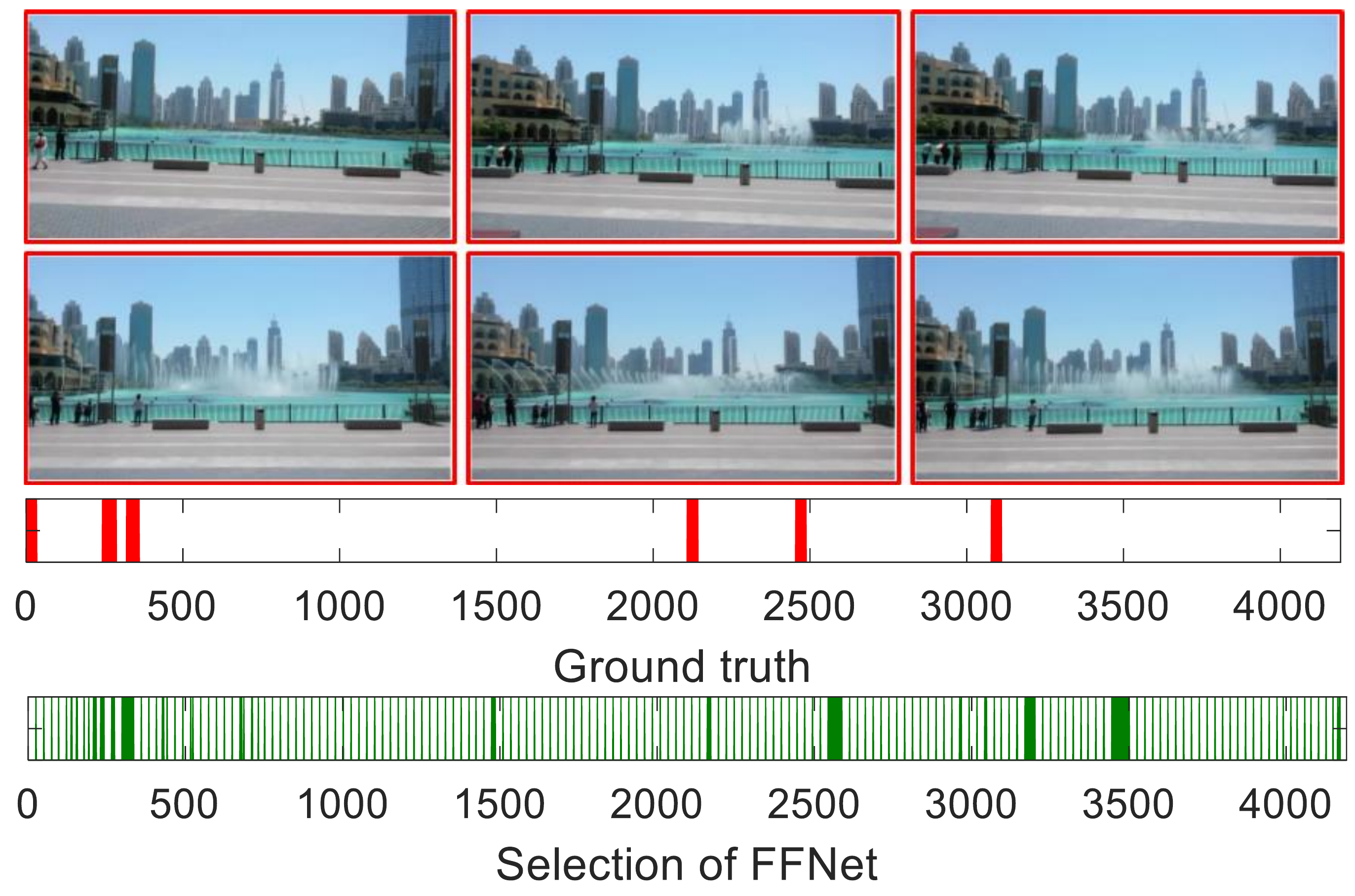}
	\end{center}
	\caption{\textbf{A failure case of FFNet.} Six frames represent the six important segments in ground truth. The ground truth selection is illustrated in the top row in red, and the selection from FFNet is illustrated below in green (figure is best viewed in color).}
	\label{failure}
\end{figure}

There are also some limitations of our model. Fig.~\ref{failure} shows a failure case of FFNet. This video records a water fountain scene near Burj Khalifa, captured by a nearly static camera. From beginning to end, the frames are almost the same, except for the change of the water fountain shape and some moving pedestrians. 
Our FFNet is able to stress on several segments, but they do not match well with the ground truth. We believe this is due to the fact that the Q agent gets similar state after each transition, which make it confused about the pattern of fast-forwarding policy for this particular video. We expect our approach could be made more robust to handle such videos by explicitly using semantic analysis~\cite{mei2013near} and could also benefit from domain adaptation techniques~\cite{kulis2011you} for more challenging datasets.

\begin{figure}
	\begin{center}
		\includegraphics[width=0.8\linewidth]{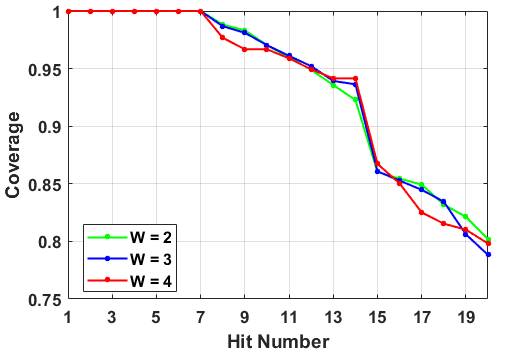}
	\end{center}
	\caption{\textbf{Effect of window size in reward.} As can be seen, it has little effect on the performance. Best viewed in color.}
	\label{window}
\end{figure}

\textbf{Effect of Window Size:} We test our approach on TVSum dataset with 3 cases of window size $w$ in $HR_k$, set to 2, 3, and 4. Fig.~\ref{window} shows that window size has little effect on the performance, indicating that our method is robust to the change in window size.

\subsection{User Study}

In addition to the above quantitative analysis, we performed a subjective evaluation study involving four human subjects to assess the quality of the selected video frames from different methods.

We choose a random subset of videos from each dataset, and run every method on them. All participants are asked to rate the overall quality of each selected subset of video frames by assigning a rating from 1 to 10, where 1 corresponding to \enquote{The selected frames are not at all informative in covering the important content from the original video} and 10 corresponding to \enquote{The selected frames are extremely informative in covering the important content from the original video}. For each video, the human rating is computed as the averaged rating from all participants (see supplementary for more details). Table~\ref{user_study} shows the average ratings for both Tour20 and TVSum datasets.  
For both datasets, consistent with the quantitative analysis results, our FFNet outperforms all other methods in covering the important content.

\begin{table}
	\begin{center}
		\begin{tabular}{|c|c|c|c|c|c|c|}
			\hline
			Dataset & OK & SC & MH & LL & SMRS & FFNet \\
			\hline\hline
			Tour20 & 7.96 & 8.18 & 8.49 & 5.28 & 4.18 & \textbf{8.70} \\
			\hline
			TVSum & 7.30 & 7.01 & 8.10 & 4.56 & 3.10 & \textbf{8.95} \\
			\hline
		\end{tabular}
	\end{center}
	\caption{\textbf{Human ratings for selected video frames from different methods.} The rating for each method is generated by averaging the ratings from all participants. Higher scores indicate better coverage of the important content. Our FFNet achieves the highest rating on both Tour20 and TVSum datasets.}
	\label{user_study}
\end{table}

\subsection{Processing Efficiency}

 All prior methods (OK, MH, LL, SC, SMRS) require processing the entire video ($100\%$). In contrast, our FFNet does not process the frames it skips over. In average, it only processes $18.67\%$ of the video frames, which could greatly improve computation efficiency, reduce resource requirement, and lower energy consumption. Note that the requirements on storage and communication are also reduced, but not as much. This is because the neighboring windows of the processed frames are also considered as important for users, and should be stored and transmitted (if needed).

\begin{figure}
	\begin{center}
		\includegraphics[width=\linewidth]{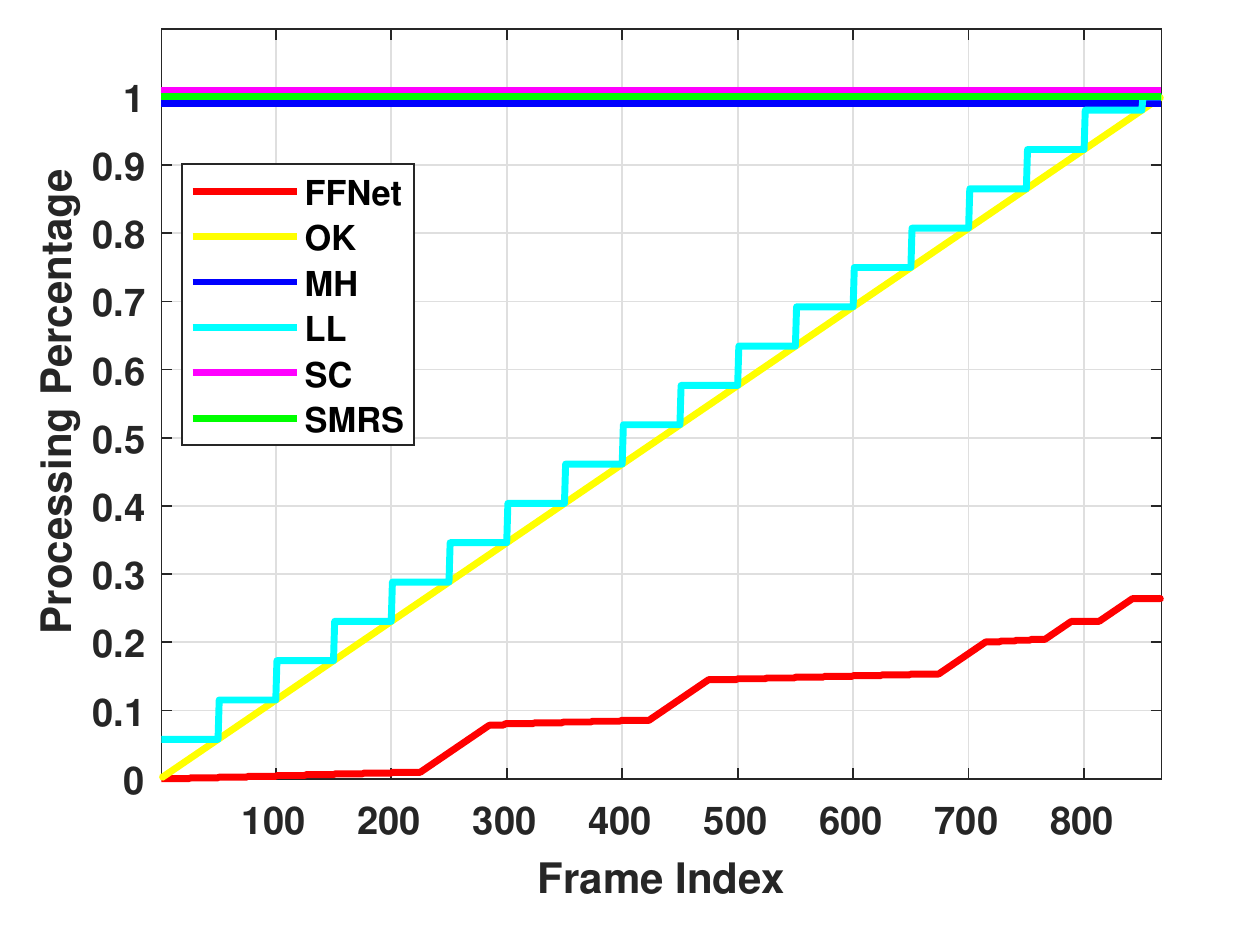}
	\end{center}
	\caption{\textbf{Example of frame processing percentage over time for different methods.} We take the video MC10 (867 frames) in Tour20 dataset as an example. Offline methods MH, SC and SMRS need to process the entire video before generating the subset. For online methods OK and LL, the processing percentage increases with time and reaches 100\% in the end. Our FFNet processes the frames based on the video content and eventually only need to process about $26\%$ of the total frames.}
	\label{rate_time}
\end{figure}

In Fig.~\ref{rate_time}, we take the video MC10 in Tour20 dataset as an example to illustrate the processing percentage over time for different methods. Microsoft Hyperlapse (MH), Spectral clustering (SC) and SMRS are offline methods that process the entire video. Online Kmeans (OK) takes frames up to the current time, and its processing percentage is linear with respect to the frame number. LiveLight (LL) updates every 50 frames, therefore the processing percentage has a step-wise shape over time. Our FFNet processes the frames at a dynamic speed based on the video content, and eventually only needs to process about $26\%$ of the total video frames.

On a Tesla K80 GPU, the average processing time per frame of FFNet is $ 8.9357*10^{-3} s$, which indicates an average frame rate of 112 fps. Most of the processing time devotes to feature extraction. The fast-forward process only takes $11.28\%$ of the time. On less-capable embedded processors, we may not achieve such high frame rate, but the low processing percentage should still help improve computation efficiency and achieving near real-time speed.

We also analyze the average running time of different methods and observe that our approach is significantly faster than the compared baselines. For a example, on a random subset videos from TVSum dataset, our proposed FFNet takes only 0.71s on to achieve 97\% coverage (at hit number 10) while the second fastest baseline SC takes 3.99s to achieve 58\% coverage and the third fastest baseline OK takes 11.58s with 71.27\% coverage.

\textbf{Supplementary Material.} Additional results and discussions along with qualitative
summaries are included in the supplementary material. We also provide details on the datasets and user study in the supplementary material.

\section{Conclusion}
In this paper, we present a supervised framework (FFNet) for fast-forwarding videos in an online fashion, by modeling the fast-forwarding operation as an Markov decision process and solving it with a Q-learning method. Quantitative and qualitative results demonstrate that FFNet outperforms multiple baseline methods in both performance and efficiency. It provides an informative subset of video frames that have better coverage of the important content in original video. At the same time, it only processes a small percentage of video frames, which improves computation efficiency and reduces requirements on various resources. 
In the future, we plan to work on integrating this method with practical system constraints like energy and available bandwidth.
It would also be interesting to extend our approach by introducing memory in the form of LSTMs--we leave this as part of the future work. 
\\
\\
\textbf{Acknowledgements:} This work was partially supported by NSF grants CNS-1544969, IIS-1724341 and CCF-1553757. The work is primarily carried out while authors Lan and Zhu were at UC Riverside.

{\small
	\bibliographystyle{ieee}
	\bibliography{egbib}

\begin{thebibliography}{10}\itemsep=-1pt

\bibitem{akyildiz2007survey}
I.~F. Akyildiz, T.~Melodia, and K.~R. Chowdhury.
\newblock A survey on wireless multimedia sensor networks.
\newblock {\em Computer networks}, 51(4):921--960, 2007.

\bibitem{arthur2007k}
D.~Arthur and S.~Vassilvitskii.
\newblock k-means++: The advantages of careful seeding.
\newblock In {\em Proceedings of the eighteenth annual ACM-SIAM symposium on
  Discrete algorithms}, 2007.

\bibitem{cheng2009smartplayer}
K.-Y. Cheng, S.-J. Luo, B.-Y. Chen, and H.-H. Chu.
\newblock Smartplayer: user-centric video fast-forwarding.
\newblock In {\em Proceedings of the SIGCHI Conference on Human Factors in
  Computing Systems}, pages 789--798, 2009.

\bibitem{elhamifar2017online}
E.~Elhamifar and M.~C. D.~P. Kaluza.
\newblock Online summarization via submodular and convex optimization.
\newblock In {\em Proceedings of the IEEE Conference on Computer Vision and
  Pattern Recognition (CVPR)}, 2017.

\bibitem{elhamifar2012see}
E.~Elhamifar, G.~Sapiro, and R.~Vidal.
\newblock See all by looking at a few: Sparse modeling for finding
  representative objects.
\newblock In {\em Proceedings of the IEEE Conference on Computer Vision and
  Pattern Recognition (CVPR)}, 2012.

\bibitem{gong2014diverse}
B.~Gong, W.~Chao, K.~Grauman, and F.~Sha.
\newblock Diverse sequential subset selection for supervised video
  summarization.
\newblock In {\em Advances in Neural Information Processing Systems (NIPS)},
  2014.

\bibitem{Top2014}
G.~Guan, Z.~Wang, S.~Mei, M.~Ott, M.~He, and D.~D. Feng.
\newblock {A Top-Down Approach for Video Summarization}.
\newblock {\em ACM Transactions on Multimedia Computing, Communications, and
  Applications}, 11(1):4, 2014.

\bibitem{gygli2014creating}
M.~Gygli, H.~Grabner, H.~Riemenschneider, and L.~Van~Gool.
\newblock Creating summaries from user videos.
\newblock In {\em European Conference on Computer Vision (ECCV)}, 2014.

\bibitem{gygli2015video}
M.~Gygli, H.~Grabner, and L.~Van~Gool.
\newblock Video summarization by learning submodular mixtures of objectives.
\newblock In {\em Proceedings of the IEEE Conference on Computer Vision and
  Pattern Recognition (CVPR)}, 2015.

\bibitem{halperin2017egosampling}
T.~Halperin, Y.~Poleg, C.~Arora, and S.~Peleg.
\newblock Egosampling: Wide view hyperlapse from egocentric videos.
\newblock {\em IEEE Transactions on Circuits and Systems for Video Technology},
  2017.

\bibitem{jiang2010new}
J.~Jiang and X.-P. Zhang.
\newblock A new player-enabled rapid video navigation method using temporal
  quantization and repeated weighted boosting search.
\newblock In {\em Computer Vision and Pattern Recognition Workshops (CVPRW),
  IEEE Computer Society Conference on}, 2010.

\bibitem{jiang2011smart}
J.~Jiang and X.-P. Zhang.
\newblock A smart video player with content-based fast-forward playback.
\newblock In {\em Proceedings of the 19th ACM international conference on
  Multimedia}, 2011.

\bibitem{joshi2015real}
N.~Joshi, W.~Kienzle, M.~Toelle, M.~Uyttendaele, and M.~F. Cohen.
\newblock Real-time hyperlapse creation via optimal frame selection.
\newblock {\em ACM Transactions on Graphics}, 34(4):63, 2015.

\bibitem{khosla2013large}
A.~Khosla, R.~Hamid, C.-J. Lin, and N.~Sundaresan.
\newblock Large-scale video summarization using web-image priors.
\newblock In {\em Proceedings of the IEEE Conference on Computer Vision and
  Pattern Recognition (CVPR)}, 2013.

\bibitem{Joint2014}
G.~Kim, L.~Sigal, and E.~P. Xing.
\newblock Joint summarization of large-scale collections of web images and
  videos for storyline reconstruction.
\newblock In {\em Proceedings of the IEEE Conference on Computer Vision and
  Pattern Recognition (CVPR)}, 2014.

\bibitem{krizhevsky2012imagenet}
A.~Krizhevsky, I.~Sutskever, and G.~E. Hinton.
\newblock Imagenet classification with deep convolutional neural networks.
\newblock In {\em Advances in neural information processing systems}, 2012.

\bibitem{Krull_2017_CVPR}
A.~Krull, E.~Brachmann, S.~Nowozin, F.~Michel, J.~Shotton, and C.~Rother.
\newblock Poseagent: Budget-constrained 6d object pose estimation via
  reinforcement learning.
\newblock In {\em Proceedings of the IEEE Conference on Computer Vision and
  Pattern Recognition (CVPR)}, 2017.

\bibitem{kulis2011you}
B.~Kulis, K.~Saenko, and T.~Darrell.
\newblock What you saw is not what you get: Domain adaptation using asymmetric
  kernel transforms.
\newblock In {\em Proceedings of the IEEE Conference on Computer Vision and
  Pattern Recognition (CVPR)}, 2011.

\bibitem{ma2013survey}
T.~Ma, M.~Hempel, D.~Peng, and H.~Sharif.
\newblock A survey of energy-efficient compression and communication techniques
  for multimedia in resource constrained systems.
\newblock {\em IEEE Communications Surveys \& Tutorials}, 15(3):963--972, 2013.

\bibitem{Att2005}
Y.~F. Ma, X.~S. Hua, and H.~J. Zhang.
\newblock A generic framework of user attention model and its application in
  video summarization.
\newblock {\em IEEE Transactions on Multimedia}, 2005.

\bibitem{Mathe_2016_CVPR}
S.~Mathe, A.~Pirinen, and C.~Sminchisescu.
\newblock Reinforcement learning for visual object detection.
\newblock In {\em Proceedings of the IEEE Conference on Computer Vision and
  Pattern Recognition (CVPR)}, 2016.

\bibitem{mei2013near}
T.~Mei, L.-X. Tang, J.~Tang, and X.-S. Hua.
\newblock Near-lossless semantic video summarization and its applications to
  video analysis.
\newblock {\em ACM Transactions on Multimedia Computing, Communications, and
  Applications}, 9(3):16, 2013.

\bibitem{mnih2015human}
V.~Mnih, K.~Kavukcuoglu, D.~Silver, A.~A. Rusu, J.~Veness, M.~G. Bellemare,
  A.~Graves, M.~Riedmiller, A.~K. Fidjeland, G.~Ostrovski, et~al.
\newblock Human-level control through deep reinforcement learning.
\newblock {\em Nature}, 518(7540):529--533, 2015.

\bibitem{money2008video}
A.~G. Money and H.~Agius.
\newblock Video summarisation: A conceptual framework and survey of the state
  of the art.
\newblock {\em Journal of Visual Communication and Image Representation},
  19(2):121--143, 2008.

\bibitem{ou2014low}
S.-H. Ou, C.-H. Lee, V.~S. Somayazulu, Y.-K. Chen, and S.-Y. Chien.
\newblock Low complexity on-line video summarization with gaussian mixture
  model based clustering.
\newblock In {\em Acoustics, Speech and Signal Processing (ICASSP), IEEE
  International Conference on}, 2014.

\bibitem{panda2017weakly}
R.~Panda, A.~Das, Z.~Wu, J.~Ernst, and A.~K. Roy-Chowdhury.
\newblock Weakly supervised summarization of web videos.
\newblock In {\em IEEE International Conference on Computer Vision (ICCV)},
  2017.

\bibitem{panda2017diversity}
R.~Panda, N.~C. Mithun, and A.~Roy-Chowdhury.
\newblock Diversity-aware multi-video summarization.
\newblock {\em IEEE Transactions on Image Processing}, 26(10):4712--4724, 2017.

\bibitem{panda2017collaborative}
R.~Panda and A.~K. Roy-Chowdhury.
\newblock Collaborative summarization of topic-related videos.
\newblock In {\em Proceedings of the IEEE Conference on Computer Vision and
  Pattern Recognition (CVPR)}, 2017.

\bibitem{peker2003extended}
K.~A. Peker, A.~Divakaran, et~al.
\newblock An extended framework for adaptive playback-based video
  summarization.
\newblock In {\em Internet Multimedia Management Systems IV}, 2003.

\bibitem{peker2001constant}
K.~A. Peker, A.~Divakaran, and H.~Sun.
\newblock Constant pace skimming and temporal sub-sampling of video using
  motion activity.
\newblock In {\em IEEE International Conference on Image Processing (ICIP)},
  2001.

\bibitem{petrovic2005adaptive}
N.~Petrovic, N.~Jojic, and T.~S. Huang.
\newblock Adaptive video fast forward.
\newblock {\em Multimedia Tools and Applications}, 26(3):327--344, 2005.

\bibitem{poleg2015egosampling}
Y.~Poleg, T.~Halperin, C.~Arora, and S.~Peleg.
\newblock Egosampling: Fast-forward and stereo for egocentric videos.
\newblock In {\em Proceedings of the IEEE Conference on Computer Vision and
  Pattern Recognition (CVPR)}, 2015.

\bibitem{Category2014}
D.~Potapov, M.~Douze, Z.~Harchaoui, and C.~Schmid.
\newblock Category-specific video summarization.
\newblock In {\em European Conference on Computer Vision (ECCV)}, 2014.

\bibitem{ramos2016fast}
W.~L. Ramos, M.~M. Silva, M.~F. Campos, and E.~R. Nascimento.
\newblock Fast-forward video based on semantic extraction.
\newblock In {\em IEEE International Conference on Image Processing (ICIP)},
  2016.

\bibitem{Ren_2017_CVPR}
Z.~Ren, X.~Wang, N.~Zhang, X.~Lv, and L.-J. Li.
\newblock Deep reinforcement learning-based image captioning with embedding
  reward.
\newblock In {\em Proceedings of the IEEE Conference on Computer Vision and
  Pattern Recognition (CVPR)}, 2017.

\bibitem{riedmiller2005neural}
M.~Riedmiller.
\newblock Neural fitted q iteration-first experiences with a data efficient
  neural reinforcement learning method.
\newblock In {\em European Conference on Machine Learning (ECML)}, 2005.

\bibitem{sigurdsson2016learning}
G.~A. Sigurdsson, X.~Chen, and A.~Gupta.
\newblock Learning visual storylines with skipping recurrent neural networks.
\newblock In {\em European Conference on Computer Vision (ECCV)}, 2016.

\bibitem{silva2016towards}
M.~M. Silva, W.~L.~S. Ramos, J.~P.~K. Ferreira, M.~F.~M. Campos, and E.~R.
  Nascimento.
\newblock Towards semantic fast-forward and stabilized egocentric videos.
\newblock In {\em European Conference on Computer Vision (ECCV)}, 2016.

\bibitem{silver2016mastering}
D.~Silver, A.~Huang, C.~J. Maddison, A.~Guez, L.~Sifre, G.~Van Den~Driessche,
  J.~Schrittwieser, I.~Antonoglou, V.~Panneershelvam, M.~Lanctot, et~al.
\newblock Mastering the game of go with deep neural networks and tree search.
\newblock {\em Nature}, 529(7587):484--489, 2016.

\bibitem{singh2014survey}
D.~Singh, G.~Tripathi, and A.~J. Jara.
\newblock A survey of internet-of-things: Future vision, architecture,
  challenges and services.
\newblock In {\em Internet of things (WF-IoT), 2014 IEEE world forum on}, 2014.

\bibitem{song2015tvsum}
Y.~Song, J.~Vallmitjana, A.~Stent, and A.~Jaimes.
\newblock Tvsum: Summarizing web videos using titles.
\newblock In {\em Proceedings of the IEEE Conference on Computer Vision and
  Pattern Recognition (CVPR)}, 2015.

\bibitem{su2016leaving}
Y.-C. Su and K.~Grauman.
\newblock Leaving some stones unturned: dynamic feature prioritization for
  activity detection in streaming video.
\newblock In {\em European Conference on Computer Vision (ECCV)}, 2016.

\bibitem{truong2007video}
B.~T. Truong and S.~Venkatesh.
\newblock Video abstraction: A systematic review and classification.
\newblock {\em ACM transactions on multimedia computing, communications, and
  applications}, 3(1):3, 2007.

\bibitem{von2007tutorial}
U.~Von~Luxburg.
\newblock A tutorial on spectral clustering.
\newblock {\em Statistics and computing}, 17(4):395--416, 2007.

\bibitem{watkins1992q}
C.~J. Watkins and P.~Dayan.
\newblock Q-learning.
\newblock {\em Machine learning}, 8(3-4):279--292, 1992.

\bibitem{wei2017deep}
T.~Wei, Y.~Wang, and Q.~Zhu.
\newblock Deep reinforcement learning for building hvac control.
\newblock In {\em Proceedings of the 54th Annual Design Automation Conference},
  2017.

\bibitem{yeung2016end}
S.~Yeung, O.~Russakovsky, G.~Mori, and L.~Fei-Fei.
\newblock End-to-end learning of action detection from frame glimpses in
  videos.
\newblock In {\em Proceedings of the IEEE Conference on Computer Vision and
  Pattern Recognition (CVPR)}, 2016.

\bibitem{Yun_2017_CVPR}
S.~Yun, J.~Choi, Y.~Yoo, K.~Yun, and J.~Young~Choi.
\newblock Action-decision networks for visual tracking with deep reinforcement
  learning.
\newblock In {\em Proceedings of the IEEE Conference on Computer Vision and
  Pattern Recognition (CVPR)}, 2017.

\bibitem{zhang2016summary}
K.~Zhang, W.-L. Chao, F.~Sha, and K.~Grauman.
\newblock Summary transfer: Exemplar-based subset selection for video
  summarizatio.
\newblock In {\em Proceedings of the IEEE Conference on Computer Vision and
  Pattern Recognition (CVPR)}, 2016.

\bibitem{zhang2016video}
K.~Zhang, W.-L. Chao, F.~Sha, and K.~Grauman.
\newblock Video summarization with long short-term memory.
\newblock In {\em European Conference on Computer Vision (ECCV)}, 2016.

\bibitem{zhao2014quasi}
B.~Zhao and E.~P. Xing.
\newblock Quasi real-time summarization for consumer videos.
\newblock In {\em Proceedings of the IEEE Conference on Computer Vision and
  Pattern Recognition (CVPR)}, 2014.

\end{thebibliography}
}

\end{document}